\documentclass[10.5pt,compsoc]{tst}
\usepackage{graphicx}
\usepackage{footmisc}
\usepackage{subfigure}
\usepackage{url}
\usepackage{multirow}
\usepackage[noadjust]{cite}
\usepackage{amsmath,amsthm}
\usepackage{amssymb,amsfonts}
\usepackage{booktabs}
\usepackage{color}
\usepackage{ccaption}
\usepackage{booktabs}
\usepackage{float}
\usepackage{fancyhdr}
\usepackage{caption}
\usepackage{xcolor,stfloats}
\usepackage{comment}
\graphicspath{{figures/}}
\usepackage{cuted}  
\usepackage{captionhack}
\usepackage{epstopdf}

\usepackage[utf8]{inputenc} 
\usepackage[T1]{fontenc}    
\usepackage{hyperref}       
\usepackage{url}            
\usepackage{booktabs}       
\usepackage{amsfonts}       
\usepackage{nicefrac}       
\usepackage{microtype}      
\usepackage{diagbox}
\usepackage{makecell}
\usepackage{enumitem}
\usepackage{multirow}
\usepackage{float}
\usepackage{mathtools}
\usepackage{graphicx}
\usepackage{soul}
\usepackage{color}
\usepackage{amsmath,bm}

\usepackage{caption}
\headevenname{\normalsize{\textbf{\emph{Tsinghua Science and Technology, xxxxx}} 20xx, 2x(x): xxx--xxx}}%
\headoddname{{\sf Ziheng Duan et al.:}\quad {\textbf{\emph{Multivariate Time Series Forecasting with Transfer Entropy Graph}}}}%

\newtheoremstyle{mystyle}{0pt}{0pt}{\normalfont}{1em}{\bf}{}{1em}{}
\theoremstyle{mystyle}
\renewcommand\figurename{Fig.~}

\newcommand{\nop}[1]{}

\makeatletter
\renewcommand{\@biblabel}[1]{[#1]\hfill}
\makeatother

\begin{document}

\hyphenpenalty=50000

\makeatletter
\newcommand\mysmall{\@setfontsize\mysmall{7}{9.5}}

\newenvironment{tablehere}
  {\def\@captype{table}}
  {}
\newenvironment{figurehere}
  {\def\@captype{figure}}
  {}

\thispagestyle{plain}%
\thispagestyle{empty}%

\let\temp\footnote
\renewcommand \footnote[1]{\temp{\normalsize #1}}
{}
\vspace*{-40pt}
\noindent{\normalsize\textbf{\scalebox{0.885}[1.0]{\makebox[5.9cm][s]
{TSINGHUA\, SCIENCE\, AND\, TECHNOLOGY}}}}

\vskip .2mm
{\normalsize
\textbf{
\hspace{-5mm}
\scalebox{1}[1.0]{\makebox[5.6cm][s]{%
I\hfill S\hfill S\hfill N\hfill{\color{white}%
l\hfill l\hfill}1\hfill0\hfill0\hfill7\hfill-\hfill0\hfill2\hfill1\hfill4
\hfill \color{white}{\quad 0\hfill ?\hfill /\hfill ?\hfill ?\quad p\hfill p\hfill  ?\hfill ?\hfill ?\hfill --\hfill ?\hfill ?\hfill ?}\hfill}}}}

\vskip .2mm
{\normalsize
\textbf{
\hspace{-5mm}
\scalebox{1}[1.0]{\makebox[5.6cm][s]{%
DOI:~\hfill~\hfill1\hfill0\hfill.\hfill2\hfill6\hfill5\hfill9\hfill9\hfill/\hfill T\hfill S\hfill T\hfill.\hfill2\hfill0\hfill 2\hfill 1\hfill.\hfill9\hfill0\hfill1\hfill0\hfill 0\hfill 8\hfill 1}}}}

\vskip .2mm\noindent
{\normalsize\textbf{\scalebox{1}[1.0]{\makebox[5.6cm][s]{%
\color{black}{V\hfill o\hfill l\hfill u\hfill m\hfill%
e\hspace{0.356em}xx,\hspace{0.356em}N\hfill u\hfill%
m\hfill b\hfill e\hfill r\hspace{0.356em}x,\hspace{0.356em}%
x\hfill x\hfill x\hfill x\hfill x\hfill%
x\hfill x\hfill \hspace{0.356em}2\hfill0\hfill x\hfill x}}}}}\\

\begin{strip}
{\center
{\LARGE\textbf{
Multivariate Time Series Forecasting with Transfer Entropy Graph}}
\vskip 9mm}

{\center {\sf \large
Ziheng Duan, Haoyan Xu, Yida Huang, Jie Feng, Yueyang Wang$^*$
}
\vskip 5mm}

\centering{
\begin{tabular}{p{160mm}}

{\normalsize
\linespread{1.6667} %
\noindent
\bf{Abstract:} {\sf
Multivariate time series (MTS) forecasting is an essential problem in many fields. Accurate forecasting results can effectively help decision-making. To date, many MTS forecasting methods have been proposed and widely applied. However, these methods assume that the predicted value of a single variable is affected by all other variables, which ignores the causal relationship among variables. To address the above issue, we propose a novel end-to-end deep learning model, termed graph neural network with Neural Granger Causality (CauGNN) in this paper. To characterize the causal information among variables, we introduce the Neural Granger Causality graph in our model. Each variable is regarded as a graph node, and each edge represents the casual relationship between variables. In addition, convolutional neural network (CNN) filters with different perception scales are used for time series feature extraction, which is used to generate the feature of each node. Finally, Graph Neural Network (GNN) is adopted to tackle the forecasting problem of graph structure generated by MTS. Three benchmark datasets from the real world are used to evaluate the proposed CauGNN. The comprehensive experiments show that the proposed method achieves state-of-the-art results in the MTS forecasting task.}
}

\end{tabular}
}
\vskip 6mm

\vskip -3mm
\small\end{strip}

\thispagestyle{plain}%
\thispagestyle{empty}%
\makeatother
\pagestyle{tstheadings}

\begin{figure}[b]
\vskip -6mm
\begin{tabular}{p{44mm}}
\toprule\\
\end{tabular}
\vskip -4.5mm
\noindent
\setlength{\tabcolsep}{1pt}
\begin{tabular}{p{1.5mm}p{79.5mm}}
$\bullet$& Ziheng Duan and Yueyang Wang are with the School of Big Data and Software Engineering, Chongqing University, Chongqing 401331, China. E-mail: duanziheng@zju.edu.cn, yueyangw@cqu.edu.cn\\
$\bullet$& Haoyan Xu, Yida Huang and Jie Feng are with the Department of Control Science and Engineering, Zhejiang University, Hangzhou 310027, China. E-mail: \{haoyanxu, stevenhuang, zjucse\_fj\}@zju.edu.cn
 \\
$\sf{*}$&
Corresponding author. \\
          &          Manuscript received: 23-Sep-2021;
          accepted: 22-Oct-2021

\end{tabular}
\end{figure}\large

\section{Introduction}
\label{s:introduction}
\noindent
In the real world, multivariate time series (MTS) data are common in various fields \cite{xu2020multivariate}, such as the sensor data in the Internet of things, the traffic flows on highways, and the prices collected from stock markets (e.g., metals price) \cite{liu2020non}.
In recent years, many time series forecasting methods have been widely studied and applied \cite{wang2020mthetgnn}.
For univariate situations, autoregressive integrated moving average model (ARIMA) \cite{box2015time} is one of the most classic forecasting methods.
However, due to the high computational complexity, ARIMA is not suitable for multivariate situations. VAR \cite{hamilton1994time, lutkepohl2005new, box2015time} method is an extended multivariate version of the AR model, which is widely used in MTS forecasting tasks due to its simplicity. But it cannot handle the nonlinear relationships among variables, which reduces its forecasting accuracy.

In addition to traditional statistical methods, deep learning methods are also applied for the MTS forecasting problem \cite{tokgoz2018rnn}.
The recurrent neural network (RNN) \cite{elman1990finding} and its two improved versions, namely the long short term memory (LSTM) \cite{hochreiter1997long} and the gated recurrent unit (GRU) \cite{chung2014empirical}, realize the extraction of time series dynamic information through the memory mechanism.
LSTNet \cite{DBLP:journals/corr/LaiCYL17} encodes short-term local information into low dimensional vectors using 1D convolutional neural networks and decodes the vectors through a recurrent neural network.
However, the existing deep learning methods cannot model the pairwise causal dependencies among MTS variables explicitly.
For example, the future traffic flow of a specific street is easier to be influenced and predicted by the traffic information of the neighboring area. In contrast, the knowledge of the area farther away is relatively useless \cite{zhou2020parallel}.
If such prior causal information can be considered, it is more conducive to the interaction among variables with causality.

Granger causality analysis (G-causality) \cite{granger1969investigating, kirchgassner2013granger} is one of the most famous studies on the quantitative characterization of time series causality.
However, as a linear model, G-causality cannot well handle nonlinear relationships.
Then transfer entropy (TE)
\cite{bossomaier2016transfer} is proposed for causal analysis, which can deal with the nonlinear cases.
TE has been widely used in the economic \cite{dimpfl2013using}, biological \cite{tung2007inferring} and industrial \cite{bauer2006finding} fields.

We propose a novel framework called Graph Neural Network with Causality (CauGNN) to further address the above limitation for MTS forecasting tasks.
After the pairwise TE among variables is calculated, the TE matrix can be obtained, regarded as the adjacency matrix of the graph structure. Each variable is one node of this graph.
In addition, CNN filters with different perception scales are used for time series feature extraction, which is used to generate the feature of each node.
Finally, Graph Neural Network (GNN) is adopted to tackle the embedding and forecasting problem of the graph generated by MTS. Our major contributions are:

\begin{itemize}[leftmargin=9pt]
\item To the best of our knowledge, we first propose an end-to-end deep learning framework that considers multivariate time series as a graph structure with causality.
\item We use transfer entropy to extract the causality among time series and construct the TE graph as a priori information to guide the forecasting task.
\item We conduct extensive experiments on MTS benchmark datasets, and the results have proved that CauGNN outperforms the state-of-the-art models.
\end{itemize}

\section{PRELIMINARIES}
\label{sec:preliminaries}
\subsection{Neural Granger Causality}
\label{sec:NGCausality}
Neural Granger is an improved Granger causality inference method.
It inherits the core idea of Granger causality, that is, if the addition of historical information of variable $i$ significantly improves the prediction accuracy of another variable $j$, then variable $i$ is the cause of variable $j$, and vice versa.
The difference is that the traditional linear Granger method uses the AR model for prediction. In contrast, the neural Granger uses deep learning and regularization to take the nonlinearity into account and avoids the computational complexity caused by pairwise calculation.

The Neural Granger network structure consists of two parts, including the variable selection module and prediction module.
The variable selection module is a fully connected layer that directly accepts historical time series as input.
The neural Granger method selects key variables by adding group Lasso regularization constraints to the weight parameters of this layer.
Group Lasso is an evolved version of Lasso regularization, which can divide constrained parameters into multiple subgroups.
If a specific group is not significant for prediction, the entire group of parameters will be assigned a zero value.
The variable selection module sets the weights connected to an input variable at different time points as a group.
If the weights of the subgroup are not zero under the regularization constraint, it means that the variable has a significant effect on prediction.
It is thus determined as the cause of the variable to be predicted. The second part of the Neural Granger is the prediction layer.
This part is not significantly different from the general prediction method.
Networks such as multilayer perceptron or LSTM can be used.
For each variable $\bm{x_i}$, a neural Granger network is established to find its cause variables.
The objective function of the network is as follows.

\begin{small}
\begin{equation}
    \min\limits_{W}\sum\limits_{t = K}^{T}(x_{it}-g_{i}(\bm{x_{(t-1):(t-K)}}))^2 + \lambda \sum\limits_{j = 1}^{p}\left\|(\bm{W_{:j}^{11}},...,\bm{W_{:j}^{1K}})\right\|_F,
\label{E1}
\end{equation}
\end{small}
where $x_{it}$ is the true value of the variable $\bm{x_i}$ at time $t$, $\bm{x_{(t-1):(t-K)}}$ is the value of all variables in $K$ lags, $g_i()$ is the function that specifies how lags from 1 to $K$ affect the future evolution of the series, $T$ is the observed time points, $p$ is the number of variables, $\lambda$ is the regularization coefficient, $\bm{W_{:j}^{11}},...,\bm{W_{:j}^{1K}}$ represents all the weight parameter connected with the $j$-th variable in the variable selection module and $\left\|\right\|_F$ is the F-norm.

\subsection{Graph Neural Network}
\label{sec:GNN}
The  concept  of  graph  neural  network  (GNN)  was  first proposed in \cite{scarselli2008graph}, which extended existing neural networks for processing the data represented in graph domains. A wide variety of graph neural network (GNN) models have been proposed in recent years \cite{wang2019heterogeneous,duan2021connecting}.
Most of these approaches fit within the framework of “neural message passing” proposed by Gilmeret et al.\cite{gilmer2017neural}. In the message passing framework, a GNN is viewed as a message passing algorithm where node representations are iteratively computed from the features of their neighbor nodes using a differentiable aggregation function \cite{ying2018hierarchical,xu2020cosimgnn,xu2021graph}. 

A separate line of work focuses on generalizing convolutions to graphs. The Graph Convolutional Networks (GCN) \cite{DBLP:journals/corr/KipfW16} could be regarded as an approximation of spectral-domain convolution of the graph signals. GCN convolutional operation could also be viewed as sampling and aggregating of the neighborhood information, such as GraphSAGE \cite{DBLP:journals/corr/HamiltonYL17} and FastGCN \cite{chen2018fastgcn}, enabling training in batches while sacrificing some time-efficiency.
Coming right after GCN, Graph Isomorphism Network (GIN) \cite{xu2018powerful} and k-GNNs \cite{morris2019weisfeiler} is developed, enabling more complex forms of aggregation. Graph Attention Networks (GAT) \cite{velivckovic2017graph} is another nontrivial direction to go under the topic of graph neural networks. It incorporates attention into propagation, attending over the neighbors via self-attention.
Recently, researchers have also applied GNN to time series forecasting problem.
For example, a correlational graph attention-based Long Short-Term Memory network (CGA-LSTM) was proposed in \cite{han2021correlational} and shows comparable performance.
This further reminds us of the superiority of the graph method in the task of MTS forecasting.

\section{Methodology}
\label{sec:methodology}
\begin{figure*}
\centering
\includegraphics[width=1\linewidth]{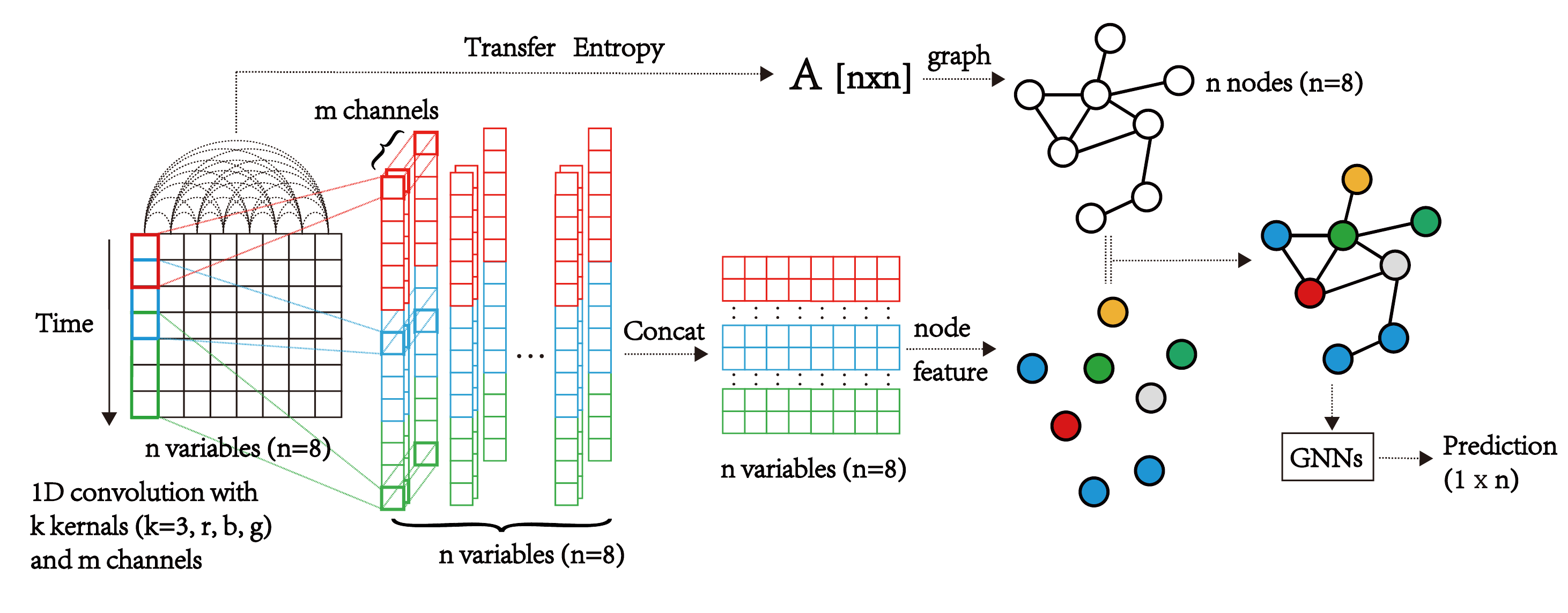}
\caption{The schematic of CauGNN. A multivariate time series consists of multiple univariate time series. CauGNN maps a multivariate time series to a graph and each univariate time series (variable) is mapped to a node. Causality matrix is calculated to model the adjacency information of nodes, while convolutional layer is used to catch node features.
The node feature matrix and adjacency matrix are then fed into graph neural network to get forecasts.}
\label{fig:general_architecture}
\end{figure*}

	
\subsection{Problem Formulation}
Given a matrix consisting of multiple observed time series $\bm{X}_n = \left[\bm{x_1}, \bm{x_2}, ..., \bm{x_t}\right]$, where $\bm{x_i}\in \mathbb{R}^n(i=1,...,n)$ and $n$ is the number of variables, the goal of MTS forecasting is to predict $\bm{x_{t+h}}$, where $h$ is the horizon ahead of the current time stamp.

\subsection{Causality Graph Structure with Transfer Entropy}
Transfer entropy (TE) is a measure of causality based on information theory, which was proposed by Schreiber in 2000 \cite{schreiber2000measuring}. 
Given a variable $\bm{X}$, its information entropy is defined as: 
\begin{equation}
    H(\bm{X})=-\sum p(x)\log_2p(x),
\end{equation}
where $x$ denotes all possible values of variable $\bm{X}$. Information entropy is used to measure the amount of information. A larger $H(\bm{X})$ indicates that the variable $\bm{X}$ contains more information. Conditional entropy is another information theory concept. Given two variables $\bm{X}$ and $\bm{Y}$, it is defined as:
\begin{equation}
    H(\bm{X}|\bm{Y})=-\sum \sum p(x,y)\log_2p(x|y),
\end{equation}
where conditional entropy $H(\bm{X}|\bm{Y})$ represents the information amount of $\bm{X}$ under the condition that the variable $\bm{Y}$ is known.
The TE of variables $\bm{Y}$ to $\bm{X}$ is defined as:

\begin{small}
\begin{eqnarray}
\begin{split}
\label{E3}
T_{\bm{Y}\to \bm{X}}
&=\sum p\Big(x_{t+1},\bm{x_t^{(k)}},\bm{y_t^{(l)}}\Big)\log_2p\Big(x_{t+1}|\bm{x_t^{(k)}},\bm{y_t^{(l)}}\Big)\! \\
& \ \ \ \ - \!\sum p\Big(x_{t+1},\bm{x_t^{(k)}}\Big)\log_2p\Big(x_{t+1}|\bm{x_t^{(k)}}\Big)\\
&=\sum p\Big(x_{t+1},\bm{x_t^{(k)}},\bm{y_t^{(l)}}\Big)\log_2\frac{p\Big(x_{t+1}|\bm{x_t^{(k)}},\bm{y_t^{(l)}}\Big)}{p\Big(x_{t+1}|{\bm{x_t^{(k)}}}\Big)}\\
&=H\Big(\bm{X_{t+1}}|\bm{X_{t}}\Big)-H\Big(\bm{X_{t+1}}|\bm{X_{t}},\bm{Y_{t}}\Big),
\end{split}
\end{eqnarray}
\end{small}
where $x_t$ and $y_t$ represent their values at time $t$. $\bm{x_t^{(k)}} = \left[x_t, x_{t-1},..., x_{t-k+1}\right]$ and $\bm{y_t^{(l)}} = \left[y_t, y_{t-1},..., y_{t-l+1}\right]$. It can be found that TE is actually an increase in the information amount of the variable $\bm{X}$ when $\bm{Y}$ changes from unknown to known. TE indicates the direction of information flow, thus characterizing causality. It is worth noting that TE is asymmetric, so the causal relationship between $\bm{X}$ and $\bm{Y}$ is usually further indicated in the following way:
\begin{equation}
\label{E4}
    T_{\bm{X},\bm{Y}}=T_{\bm{X}\rightarrow \bm{Y}}-T_{\bm{Y}\rightarrow \bm{X}}.
\end{equation}
When $T_{\bm{X},\bm{Y}}$ is greater than $0$, it means that $\bm{X}$ is the cause of $\bm{Y}$, otherwise $\bm{X}$ is the consequence of $\bm{Y}$.
In this paper, we use neural granger to characterize the causal relationship among variables. 
The causality matrix $\bm{T}$ of the multivariate time series $\bm{X}_n$ can be formulated with the element $t_{ij}$ corresponding to the $i$-th row and $j$-th column as:
\begin{equation}
t_{ij}=\left\{
        \begin{array}{lr}
             T_{\bm{x_i},\bm{x_j}}, & T_{\bm{x_i},\bm{x_j}}\textgreater c\\
             0, & otherwise,\\
        \end{array}
\right.
\end{equation}
where $\bm{x_i}$ is the $i$-th variable of $\bm{X}_n$, $c$ is the threshold to determine whether the causality is significant. $\bm{T}$ can be regarded as the adjacency matrix of the MTS graph structure.

\subsection{Feature Extraction of Multiple Receptive Fields}
\label{sec:CNN}
Time series is a special kind of data. 
When analyzing time series, it is necessary to consider not only its numerical value but also its trend over time. 
In addition, time series from the real world often have multiple meaningful periods.
For example, the traffic flow of a certain street not only shows a similar trend every day, but meaningful rules can also be observed in the unit of a week. 
Therefore, it is reasonable to extract the features of time series in units of multiple certain periods. 
In this paper, we use multiple CNN filters with different receptive fields, namely kernel sizes, to extract features at multiple time scales. 
Given an input time series $\bm{x}$ and $p$ CNN filters, denoted as $\bm{W_i}$, with different convolution kernel sizes $(1 \times k_i) (i = 1,2...p)$ are separately generated and the features $\bm{h}$ are extracted as follows: $\bm{h_i}=ReLU(\bm{W_i}*\bm{x}+\bm{b_i})$, $\bm{h}=[\bm{h_1} \oplus \bm{h_2} \oplus ... \oplus \bm{h_p}]$.
$*$ denotes the convolution operation, $[\oplus]$ represents the concatenate operation, and $ReLU$ is a nonlinear activation function $ReLU(x) = max(0,x)$. 

\subsection{Node Embedding Based on Causality Matrix}
After feature extraction, the input MTS is converted into a feature matrix $\bm{H}\in \mathbb{R}^{n\times d}$ , where $d$ is the number of features after the calculation introduced in Section \ref{sec:CNN}. 
$\bm{H}$ can be regarded as a feature matrix of a graph with $n$ nodes. 
The adjacency of nodes in the graph structure is determined by the causality matrix $\bm{T}$. 
For such graph structure, graph neural networks can be directly applied for the embedding of nodes. 
Inspired by k-GNNs \cite{morris2019weisfeiler} model, we propose CauGNN model and use the following propagation mechanism for calculating the forward-pass update of a node denoted by $\bm{v_i}$:
\begin{equation}
\bm{h_i^{(l+1)}} = \sigma\Big( \bm{h_i^{(l)}}\bm{W_1^{(l)}}+ \sum_{j\in \bm{N(i)}}\bm{h_j^{(l)}}\bm{W_2^{(l)}} \Big),
\end{equation}
where 
$\bm{W_1^{(l)}}$ and $\bm{W_2^{(l)}}$ are parameter matrices, 
$\bm{h_i^{(l)}}$ is the hidden state of node $\bm{v_i}$ in the $l^{th}$ layer and $\bm{N(i)}$ denotes the neighbors of node $i$.
k-GNNs only perform information fusion between a certain node and its neighbors, ignoring the information of other non-neighbor nodes. 
This design highlights the relationship among variables, which can effectively avoid the information redundancy brought by high dimensions. 
By adding the priori causal information obtained by TE, the model does not need to find out the key variables for forecasting by itself. 
In this paper, the output dimension of the last graph neural network layer is 1, which is used as the prediction result.

Overall, we use $\ell_1$-norm loss to measure the prediction of $\bm{x_{t+h}}$ and optimize the model via Adam algorithm \cite{kingma2014adam}.

\section{Experiments}
\label{experiments}
In this section, we conduct extensive experiments on three benchmark datasets for multivariate time series forecasting tasks, and compare the results of proposed CauGNN model with other $6$ baselines. All the data and experiment codes are available online\footnote{https://github.com/RRRussell/CauGNN.}.
\subsection{Data}
We use three benchmark datasets which are publicly available.

\begin{itemize}[leftmargin=9pt]
    \item \textbf{Exchange-Rate}\footnote{https://github.com/laiguokun/multivariate-time-series-data}: the exchange rates of eight foreign countries collected from $1990$ to $2016$, collected per day.
	\item \textbf{Energy} \cite{candanedo2017data}: measurements of $26$ different quantities related to appliances energy consumption in a single house for $4.5$ months, collected per $10$ minutes.
	\item \textbf{Nasdaq} \cite{qin2017dual}: the stock prices are selected as the multivariable time series for $82$ corporations, collected per minute.
\end{itemize}

\subsection{Methods for Comparison}
The methods in our comparative evaluation are as follows:
\begin{itemize}[leftmargin=9pt]
    \item \textbf{VAR} \cite{hamilton1994time, lutkepohl2005new, box2015time} stands for the well-known vector regression model, which has proven to be a useful machine learning method for multivariate time series forecasting.
    \item \textbf{CNN-AR} \cite{lecun1995convolutional} stands for classical convolution neural network. We use multi-layer CNN with AR components to perform MTS forecasting tasks.
    \item \textbf{RNN-GRU} \cite{chung2014empirical} is the Recurrent Neural Network using GRU cell with AR components. 
    \item \textbf{MultiHead Attention} \cite{NIPS2017_7181} stands for multihead attention components in the famous Transformer model, where multi-head mechanism runs through the scaled dot-product attention multiple times in parallel.
    \item \textbf{LSTNet} \cite{DBLP:journals/corr/LaiCYL17} is a famous MTS forecasting framework which shows great performance by  modeling long- and short-term temporal patterns of MTS data.
    \item \textbf{MLCNN} \cite{cheng2019towards} is a novel multi-task deep learning framework which adopts the idea of fusing foreacasting information of different future time.
    \item \textbf{CauGNN} stands for our proposed Graph Neural Network with Transfer Entropy. We apply multi-layer CNN and k-GNNs to perform MTS forecasting tasks.
    \item \textbf{CauGIN} stands for our proposed Graph Isomorphism Network with Transfer Entropy, where k-GNNs layers are replaced by GIN layers.
    \item \textbf{CauGNN-nCau} We remove the Transfer entropy matrix use all-one adjacency matrix instead.
    \item \textbf{CauGNN-nCNN} We remove the CNN component and use input time series data as node features.
\end{itemize}

\subsection{Metrics}
We apply three conventional evaluation metrics to evaluate the performance of different models for multivariate time series prediction: Mean Absolute Error (\textbf{MAE}), Relative Absolute Error (\textbf{RAE}), Empirical Correlation Coefficient (\textbf{CORR}):
\begin{center}
\begin{equation}
\begin{aligned}
MAE=\frac{1}{n}\sum\limits_{i = 1}^{n} |p_i-a_i|, RAE=\frac{\sum\limits_{i = 1}^{n} \left |p_i-a_i \right |}{\sum\limits_{i = 1}^{n} \left |\bar{a}-a_i \right |}s, \\
CORR=\frac{\sum\limits_{i = 1}^{n} (p_i-\bar{p})(a_i-\bar{a})}{\sqrt{\sum\limits_{i = 1}^{n} (p_i-\bar{p})^2}\sqrt{\sum\limits_{i = 1}^{n} (a_i-\bar{a})^2}}.\\
\end{aligned}
\end{equation}
   \textit{a = actual target, p = predict target}
\end{center}

For \textbf{MAE} and \textbf{RAE} metrics, lower value is better; for \textbf{CORR} metric, higher value is better.

\begin{table*}[!t]
\caption{MTS forecasting results measured by MAE/RAE/CORR score over three datasets.}
\vskip 3mm
\centering
\scalebox{1}{
\begin{tabular}{lc|ccc|ccc|ccc} 
\toprule 
    Dataset&&\multicolumn{3}{c|}{Exchange-Rate}& \multicolumn{3}{c|}{Energy} & \multicolumn{3}{c}{Nasdaq} \\
\midrule
    &&\multicolumn{1}{c}{horizon}&\multicolumn{1}{c}{horizon}&\multicolumn{1}{c|}{horizon}&\multicolumn{1}{c}{horizon}&\multicolumn{1}{c}{horizon}&\multicolumn{1}{c|}{horizon}&\multicolumn{1}{c}{horizon}&\multicolumn{1}{c}{horizon}&\multicolumn{1}{c}{horizon}\\
    Methods&Metrics&\multicolumn{1}{c}{5} &\multicolumn{1}{c}{10} &\multicolumn{1}{c|}{15} &\multicolumn{1}{c}{5} &\multicolumn{1}{c}{10}&\multicolumn{1}{c|}{15}&\multicolumn{1}{c}{5} &\multicolumn{1}{c}{10} &\multicolumn{1}{c}{15}\\

\midrule 
\multirow{3}{*}{\textsc{VAR}}
    &MAE& 0.0065 & 0.0093 & 0.0116 & 3.1628 & 4.2154 & 5.1539 & 0.1706 & 0.2667 & 0.3909\\
    &RAE& 0.0188 & 0.0270 & 0.0339 & 0.0545 & 0.0727 & 0.0889 & 0.0011 & 0.0018 & 0.0026\\
    &CORR& 0.9619 & 0.9470 & 0.9318 & 0.9106 & 0.8482 & 0.7919 & 0.9911 & 0.9273 & 0.5528\\
\midrule 

\multirow{3}{*}{\textsc{CNN-AR}}
    &MAE& 0.0063 & 0.0085 & 0.0106 & 2.4286 & 2.9499 & 3.5719 & 0.2110 & 0.2650 & 0.2663\\
    &RAE& 0.0182 & 0.0249 & 0.0303 & 0.0419 & 0.0509 & 0.0616 & 0.0014 & 0.0017 & 0.0017\\
    &CORR& 0.9638 & 0.9490 & 0.9372 & 0.9159 & 0.8618 & 0.8150 & 0.9920 & 0.9919 & 0.9860\\
\midrule 
\multirow{3}{*}{\textsc{RNN-GRU}}
    &MAE& 0.0066 & 0.0092 & 0.0122 & 2.7306 & 3.0590 & 3.7150& 0.2245 & 0.2313 &  0.2700\\
    &RAE& 0.0192 & 0.0268 & 0.0355 & 0.0471 & 0.0528 & 0.0641 & 0.0015 & 0.0015 &  0.0018\\
    &CORR& 0.9630 & 0.9491 & 0.9323 & 0.9167 & 0.8624 & 0.8106 & 0.9930 & 0.9901 &  0.9877\\
\midrule 
\multirow{3}{*}{\textsc{MultiHead Att}}
    &MAE & 0.0078 & 0.0101 & 0.0119 & 2.6155 & 3.2763 & 3.8457& 0.2218 & 0.2446 & 0.3177 \\
    &RAE&0.0227 & 0.0294 & 0.0347 & 0.0451 & 0.0565 & 0.0663& 0.0014 & 0.0017 & 0.0027 \\
    &CORR&0.9630 & 0.9500 & 0.9376 & 0.9178 & 0.8574 & 0.8106& 0.9945 & 0.9915 & 0.9857 \\
\midrule 
\multirow{3}{*}{\textsc{LSTNet}}
    &MAE& 0.0063 & 0.0085 & 0.0107 & 2.2813 & 3.0951 & 3.4979 & 0.1708 & 0.2511 & 0.2603\\
    &RAE& 0.0184 & 0.0247 & 0.0311 & 0.0393 & 0.0534 & 0.0603 & 0.0011 & 0.0016 & 0.0017\\
    &CORR& 0.9639 & 0.9490 & 0.9373 & 0.9190 & 0.8640 & 0.8216 & 0.9940 & 0.9902 & 0.9872\\
\midrule 
\multirow{3}{*}{\textsc{MLCNN}}
    &MAE& 0.0065 & 0.0094 & 0.0107 & 2.4529 & 3.4381 & 3.7557 & 0.1301 & 0.2054 & 0.2375\\
    &RAE& 0.0189 & 0.0274 & 0.0312 & 0.0423 & 0.0593 & 0.0648 & 0.0009 & 0.0013 & 0.0016\\
    &CORR&  0.9693 & \textbf{0.9559} & \textbf{0.9511} & 0.9212 & 0.8603 & 0.8121 & 0.9965 & 0.9931 & 0.9898\\
\toprule[1.2pt] 

\multirow{3}{*}{\textsc{CauGNN-nCau}}
    &MAE& 0.0076 & 0.0093 & 0.0113 & 2.1753 & 2.8731 & 3.4122 & 0.1601 & 0.2174 & 0.2490\\
    &RAE& 0.0221 & 0.0290 & 0.0315 & 0.0369 & 0.0475 & 0.0588 & 0.0010 & 0.0014 & 0.0016\\
    &CORR& 0.9660 & 0.9531 & 0.9425 & 0.9210 & 0.8587 & 0.8167 & 0.9942 & 0.9907 & 0.9879\\
\midrule 
\multirow{3}{*}{\textsc{CauGNN-nCNN}}
    &MAE& 0.0074 & 0.0096 & 0.0118 & 2.2346 & 2.7488 & 3.5229 & 0.1884 & 0.4454 & 0.3342\\
    &RAE& 0.0240 & 0.0350 & 0.0325 & 0.0575 & 0.0574 & 0.0673 & 0.0012 & 0.0029 & 0.0022\\
    &CORR& 0.9634 & 0.9518 & 0.9398 & 0.9196 & 0.8608 & 0.8121 & 0.9937 & 0.9909 & 0.9856\\
\midrule 
\multirow{3}{*}{\textsc{CauGNN}}
    &MAE& \textbf{0.0060} & \textbf{0.0083} & \textbf{0.0104} & \textbf{2.0454} & \textbf{2.7242} & \textbf{3.3232}& 0.1549 & 0.1897 & 0.2358\\
    &RAE&\textbf{0.0176} & \textbf{0.0243} & \textbf{0.0302} & \textbf{0.0358} & \textbf{0.0470} & \textbf{0.0573}& 0.0010 & 0.0012 & 0.0015\\
    &CORR& \textbf{0.9694} & 0.9548 & 0.9438 & \textbf{0.9267} & \textbf{0.8673} & \textbf{0.8221}& 0.9951 & 0.9922 & 0.9887\\
\midrule 

\multirow{3}{*}{\textsc{CauGIN}}
 	  &MAE& 0.0065 & 0.0089 & 0.0108 & 2.1768 & 2.8097 & 3.3572 & \textbf{0.1174} & \textbf{0.1664} & \textbf{0.2043}\\
    &RAE& 0.0188 & 0.0259 & 0.0315 & 0.0375 & 0.0485 & 0.0579 & \textbf{0.0008} & \textbf{0.0011} & \textbf{0.0013}\\
    &CORR& 0.9690 & 0.9551 & 0.9441 & 0.9204 & 0.8615 & 0.8131 & \textbf{0.9968} & \textbf{0.9937} & \textbf{0.9907}\\
    
\bottomrule 
\end{tabular}}

\label{sec:table1}
\end{table*}

\subsection{Experiment Details}
We conduct grid search on tunable hyper-parameters on each method over all datasets. Specifically, we set the same grid search range of input window size for each method from \{$2^0$,$2^1$,...,$2^9$\} if applied. We vary hyper-parameters for each baseline method to achieve their best performance on this task. For RNN-GRU and LSTNet, the hidden dimension of Recurrent and Convolutional layer is chosen from $\{10,20,...,100\}$. For LSTNet, the skip-length $p$ is chosen from $\{0,12,...,48\}$. For MLCNN, the hidden dimension of Recurrent and Convolutional layer is chosen from $\{10,25,50,100\}$. We adopt dropout layer after each layer, and the dropout rate is set from $\{0.1,0.2\}$. We calculate transfer entropy matrix based on train and validation data. For \textsc{CauGNN}, \textsc{CauGIN}, CauGNN-nCau, CauGNN-nCNN, we set the size of the three convolutional kernels to be $\{3,5,7\}$ respectively and the number of channels of each kernel is $12$ in all our models. The hidden dimension of k-GNNs layer is chosen from $\{10,20,...,100\}$. For \textsc{CauGIN}, the hidden size is chosen from $\{10,20,...,100\}$.
For the hyperparameter $c$, which is the threshold to determine  whether the causality is significant, we search it in the range of [0,0.1], and choose 0.005.
The Adam algorithm is used to optimize the parameters of our model.
For more details, please refer to our code.

\begin{figure*}
\centering
\includegraphics[width=1\linewidth]{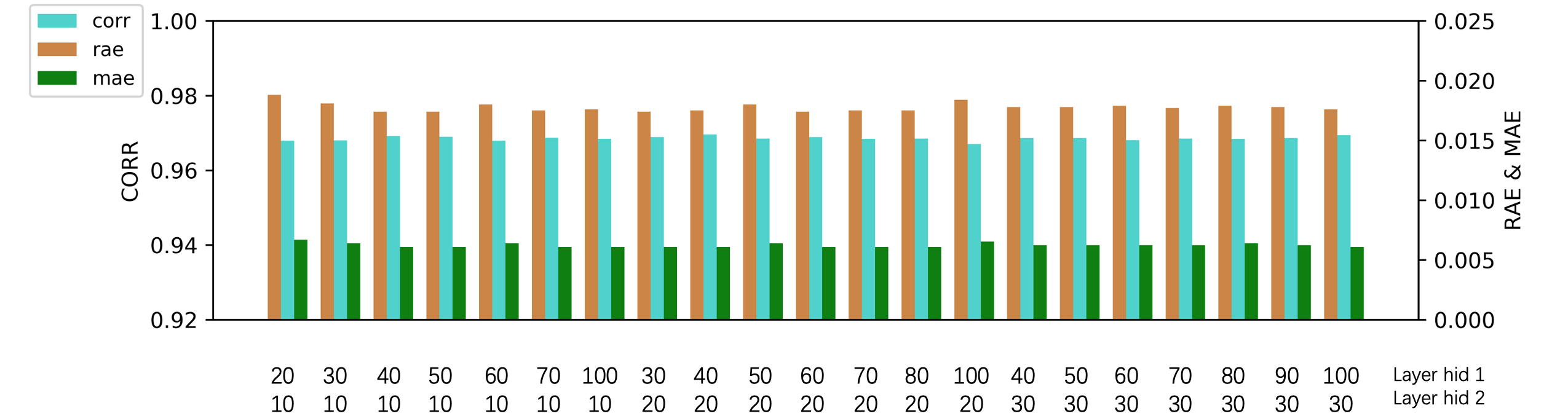}
\caption{Parameter sensitivity test results. CauGNN shows steady performance under different settings of hidden sizes in GNN layer.}
\label{fig:parameter_sensitivity}
\end{figure*}

\begin{figure}
\centering
\includegraphics[width=1\linewidth]{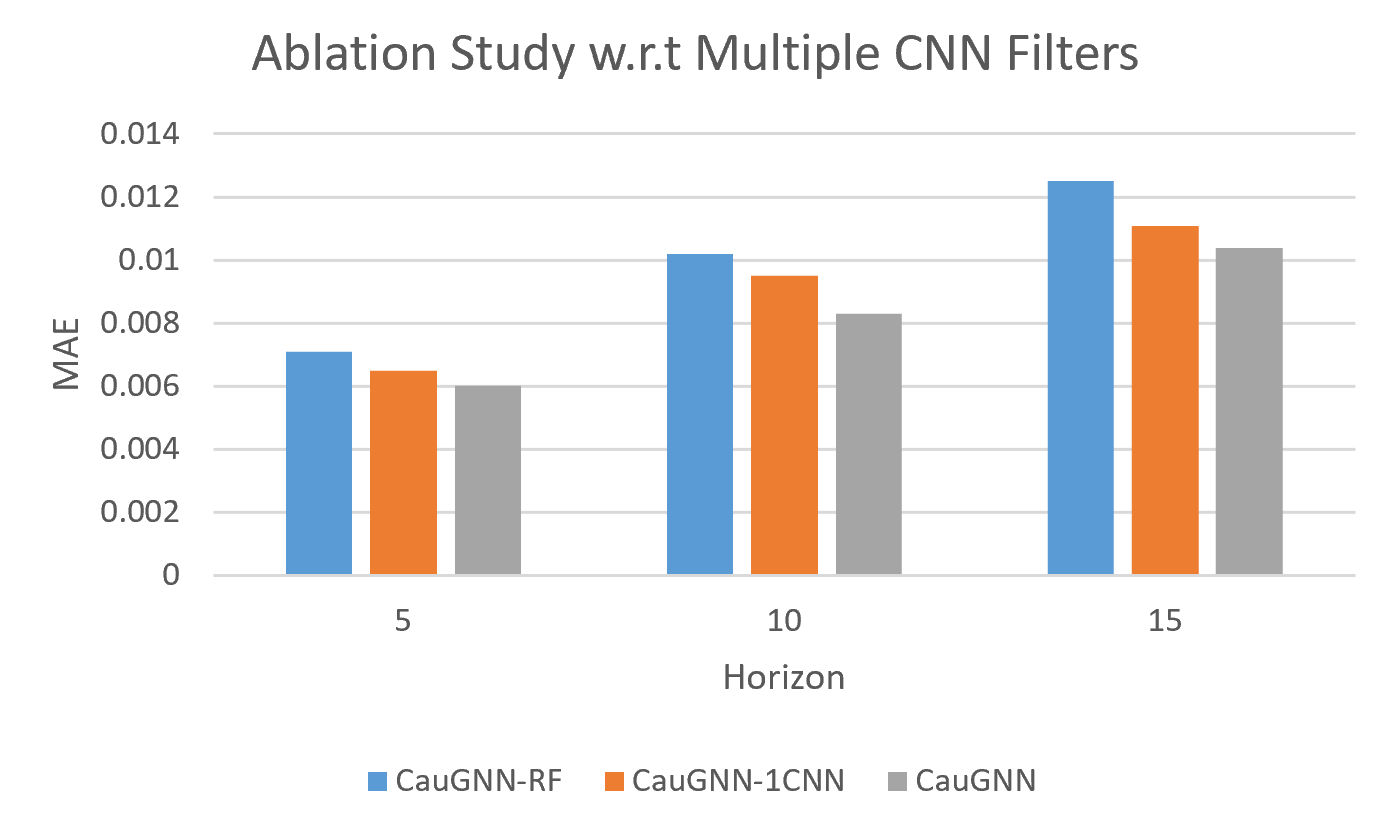}
\caption{Ablation study w.r.t. multiple CNN filters on the Exchange-Rate dataset when horizon is 5.}
\label{fig:ab}
\end{figure}

\subsection{Main Results}
Table \ref{sec:table1} summarizes the evaluation results of all the methods on 3 benchmark datasets with $3$ metrics. Following the test settings of \cite{DBLP:journals/corr/LaiCYL17}, we use each model for time series predicting on future moment $\{t+5,t+10,t+15\}$, thus we set \textit{horizon = } $\{5,10,15\}$, which means the horizon is set from $5$ to $15$ days for forecasting over the Exchange-Rate data, from $50$ to $150$ minutes over the Energy data, and from $5$ to $15$ minutes over the Nasdaq data. The best results for each metrics on each dataset is set bold in the Table .
We save the model that has the best performance on validation set based on RAE or MAE metric after training $1000$ epochs for each method. Then we use the model to test and record the results. 
%
The results shows the proposed \textsc{CauGNN} model outperforms most of the baselines in most cases, indicating the effectiveness of our proposed model on multivariate time series predicting tasks adopting the idea of using causality as guideline for forecasting. On the other side, we observe the result of VAR model on Nasdaq dataset is far worse than other methods in some cases, partly because VAR is not sensitive to the scale of input data which lower its performance.

MLCNN shows impressing results because it can fuse near and distant future visions, while LSTNet model shows impressing results when modeling periodic dependency patterns occurred in data. Our proposed \textsc{CauGNN} uses transfer entropy matrix to collect the internal relationship between variables and analyze the topology composed of variables and relationships through graph network, thus it can break through these restrictions and perform well on general datasets.  
		
Other deep learning baseline models show similar performance. This results from the fine-tuned work on general deep learning methods and the suitable hyper-parameters we used. We use the following sets of hyperparameters for RNN-GRU, MultiHead Attention, LSTNet and MLCNN: $50$ (hidCNN), $50$ (hidRNN), $5$ (hidSkip), $128$ (windowsize); RNN-GRU: $50$ (hidRNN), $24$ (highway window) on Exchange-Rate dataset, and fine-tuned adjustment over other datasets. \textsc{CauGNN} model sets $12$ (hidCNN), $30$ (hidGNN1), $10$ (hidGNN2), $32$ (window size) applying to all datasets and horizons. Compared with these baseline models, our proposed \textsc{CauGNN} model can share the same hyper-parameters among varies datasets and situations with robust performance as the results show.

\subsection{Variant Comparison}
Our proposed framework has strong universality and compatibility. We replace the k-GNNs layer with GIN layer, which also well preserves the distinctness of inputs. As showned in Table\ref{sec:table1}, GIN layer fits into our model well and \textsc{CauGIN} has similar performance with \textsc{CauGNN}.

For ablation study, we also replace transfer entropy matrix with all-one matrix in CauGNN-nCau, assuming the value to be predicted of a single variable is related to all other variables, thus a completed graph is fed into GNN layers. The experiment results show that CauGNN outperforms CauGNN-nCau, which indicates the significant role TE matrix plays in CauGNN model. On the other hand, we conduct experiments by using CauGNN-nCNN model, in which CNN component is removed. The input time series data without feature extraction are fed into GNN layer instead of node features extracted from CNN layer. The experiment results show that CauGNN outperforms CauGNN-nCNN, which suggest the significant role CNN component plays in CauGNN model.

To test the parameter sensitivity of our model, we evaluate how the hidden size of the GNN component can affect the results. We report the \textbf{MAE}, \textbf{RAE}, \textbf{CORR} metrics on Exchange-Rate dataset. As can be seen in Figure \ref{fig:parameter_sensitivity}, while ranging the hidden size of GNN layers from $\{10, 20, ..., 100\}$, the model performance is steady, being relatively insensitive to the hidden dimension parameter. 

To prove the superiority of multiple CNN filters, we also did an ablation study on the Exchange-Rate dataset when horizon is 5.
As shown in the Figure \ref{fig:ab}, CauGNN-RF, CauGNN-1CNN, and CauGNN respectively represent the direct use of the raw feature (original data) as the input of the node embedding model, using one CNN filter (here we set kernel size to 3), and our complete model CauGNN with three CNN filters.
We can find that CauGNN-RF has the worst performance, indicating that direct use of raw feature will introduce too much noise, which is not conducive to the subsequent learning of the model; and CauGNN has the best performance, indicating that stacking multiple CNN filters can better capture multiple inherent time series period characteristics and make more accurate predictions.

\section{Conclusion}
\label{sec:conclusion}
In this paper, we propose a novel deep learning framework (CauGNN) for multivariate time series forecasting. 
Using CNN with multiple receiving fields, our model introduces causal prior information with transfer entropy features and uses graph neural network for feature extraction, which effectively improves the results in MTS forecasting.
With in-depth theoretical analysis and experimental verification, we confirm that CauGNN successfully captures the causal relationship among variables and uses graph neural network to select key variables for accurate forecasting.

{ 
\bibliographystyle{unsrt}
\bibliography{saliencysegmentation}
}

\begin{biography}[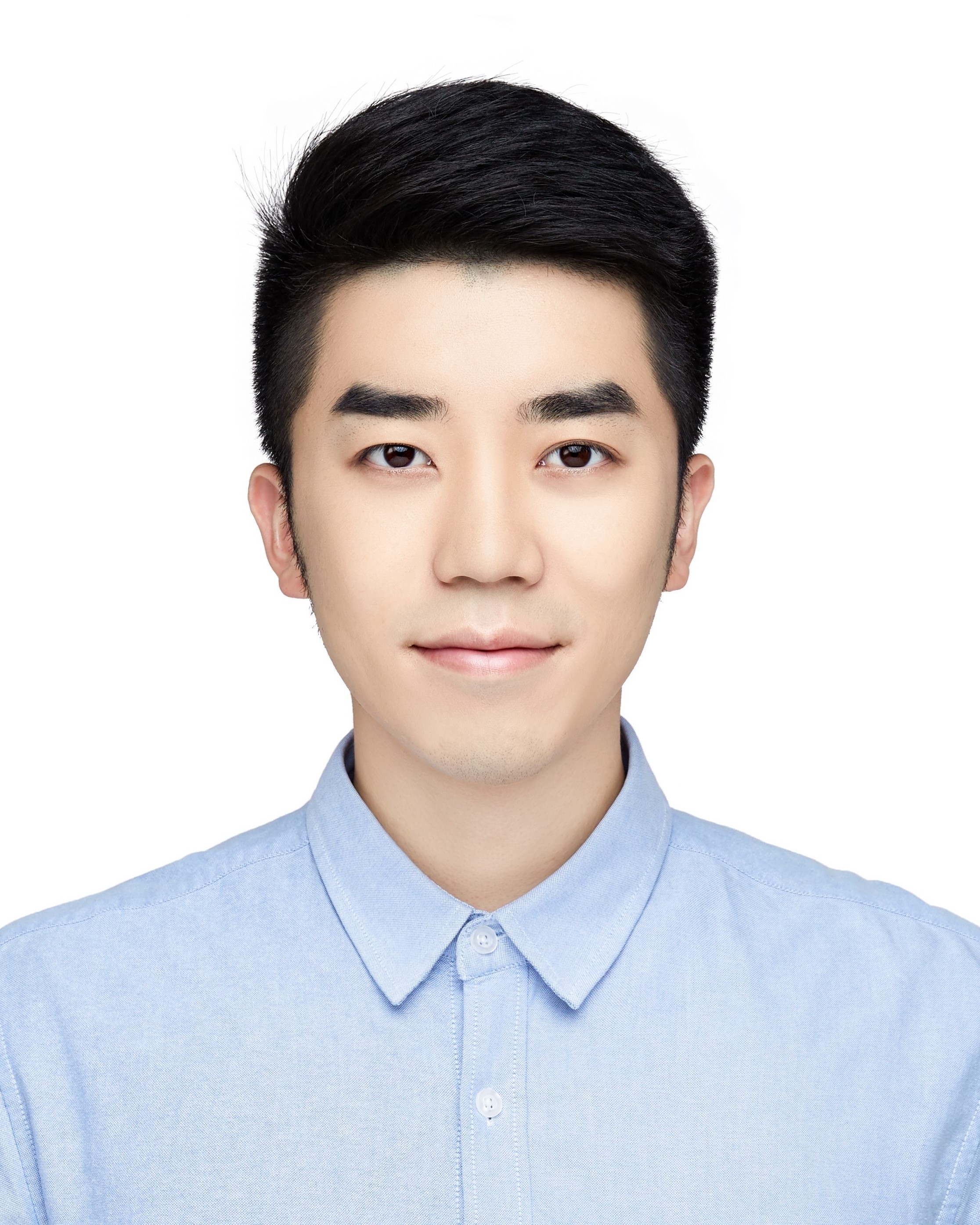]
\noindent
\textbf{Ziheng Duan}\ \  received his B.E. in July, 2020 at Zhejiang University, College of Control Science and Engineering. His research interests lie in the area of Machine Learning, Computational Biology, Graph Representation Learning, Time Series Analysis, espeically the interaction of them.
\end{biography}

\vskip 22mm
\begin{biography}[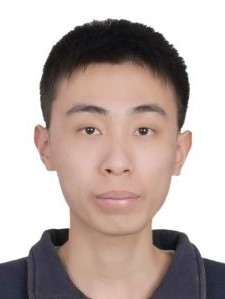]
\noindent
\textbf{Haoyan Xu} received his B.E. in July, 2020 at Zhejiang University, College of Control Science and Engineering. His research interests lie in the area of Graph Representation Learning, Time Series Analysis, Robot Learning and Microfluidics. He is particular interested in graph neural networks, with their applications in language processing, graph mining, etc.
\end{biography}

\vskip 1mm
\begin{biography}[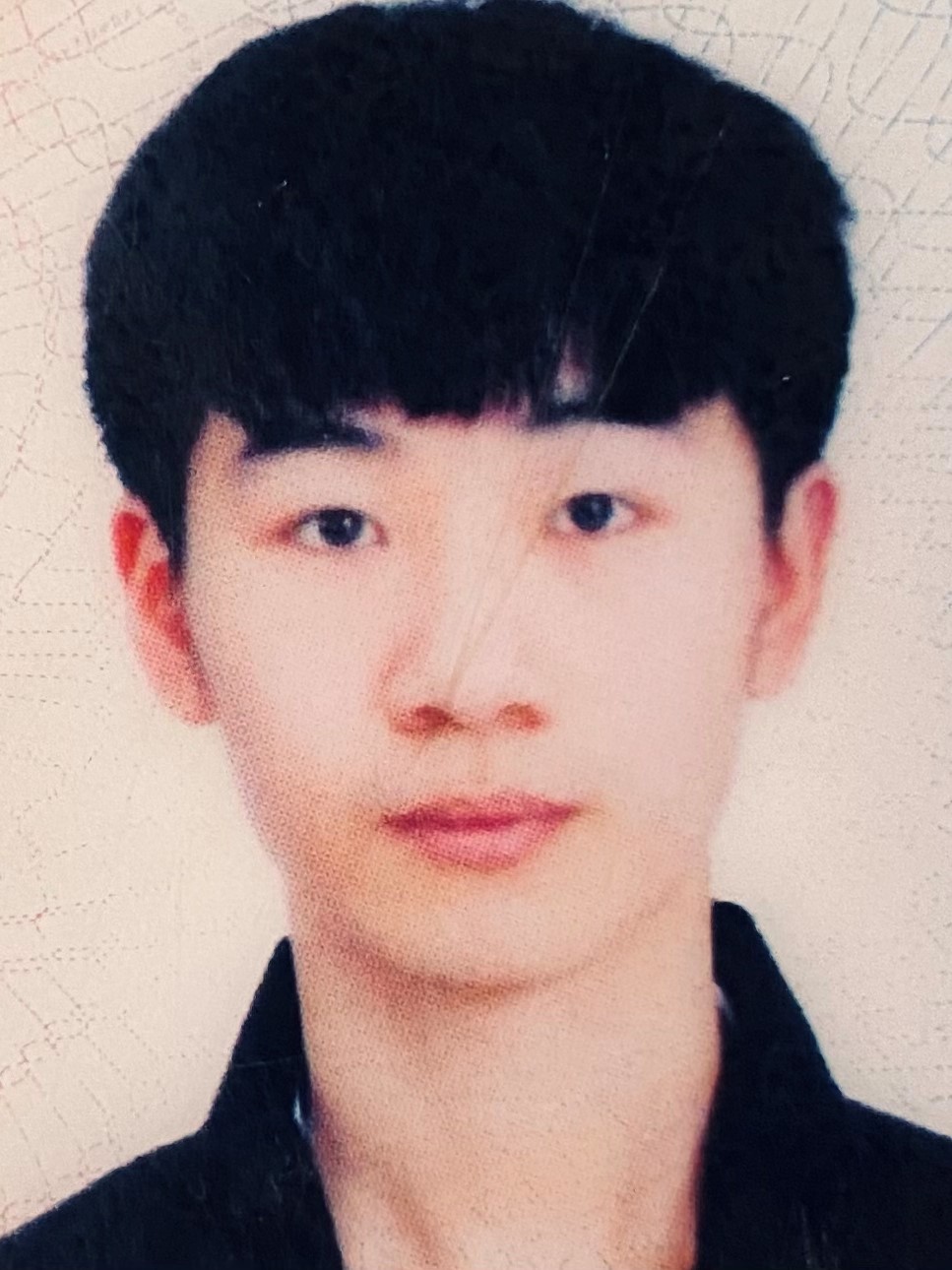]
\noindent
\textbf{Yida Huang}  received his B.E. in July, 2020 at Zhejiang University, College of Computer Science and Engineering. His research interests lie in the area of Time Series Analysis, Graph Representation Learning, Natural Language Processing, and their applications in Cyber Security.
\end{biography}

\vskip 1mm
\begin{biography}[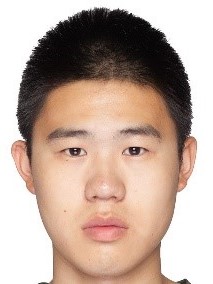]
\noindent
\textbf{Jie Feng}  received his B.E. in July, 2021 at Zhejiang University, College of Control Science and Engineering, who will receive his B.S. in June. 2021. His research interests include artificial intelligence and robotics.
\end{biography}

\vskip 1mm
\begin{biography}[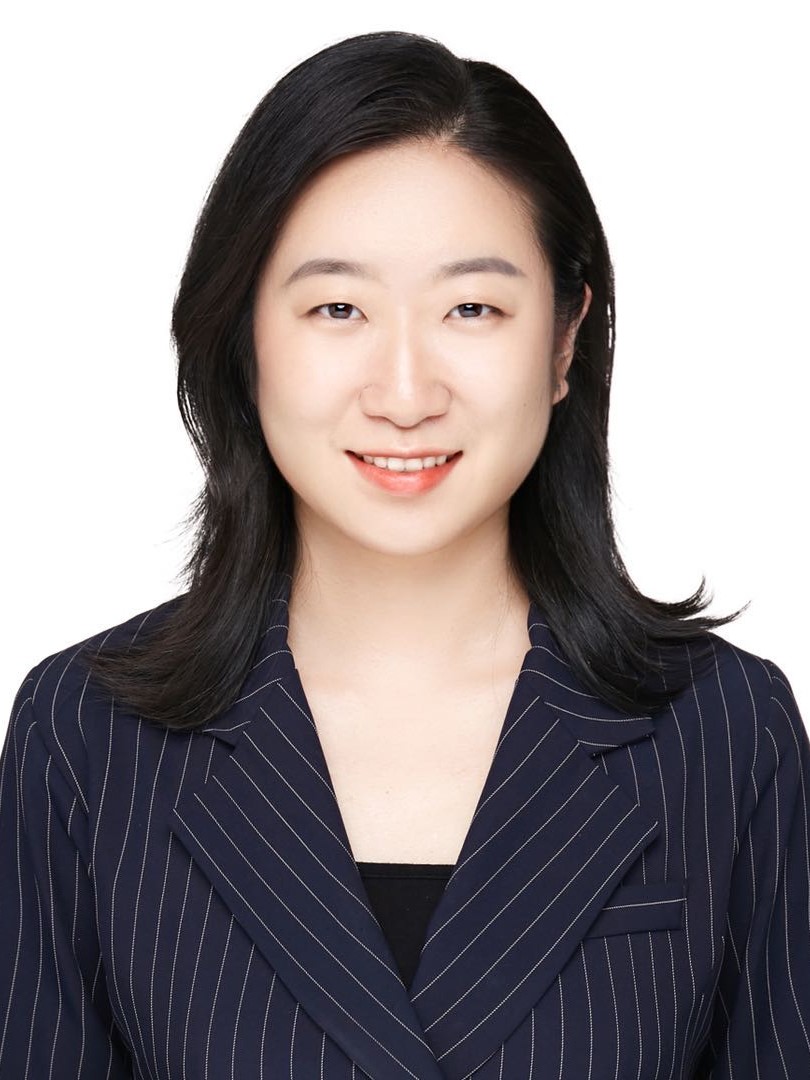]
\noindent
\textbf{Yueyang Wang} received the B.E. degree from the Software Institute, Nanjing University, Nanjing, China, and the Ph.D. degree from Zhejiang University, Hangzhou, China. She is currently a Lecturer with the School of Big Data and Software Engineering, Chongqing University. Her research interests include social network analysis and data mining.
\end{biography}

\end{document}